# Flatten-T Swish: a thresholded ReLU-Swish-like activation function for deep learning

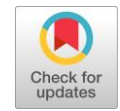

Hock Hung Chieng [a,1,*], Noorhaniza Wahid [a,2], Ong Pauline [b,3], Sai Raj Kishore Perla [c,3]

[a] Faculty of Computer Science and Information Technology, Universiti Tun Hussein Onn Malaysia, Johor, Malaysia
[b] Faculty of Mechanical and Manufacturing Engineering, Universiti Tun Hussein Onn Malaysia, Johor, Malaysia
[c] Department of Electronics and Communication Engineering, Institute of Engineering and Management, Kolkata, India
[1] hi160029@siswa.uthm.edu.my; [2] nhaniza@uthm.edu.my; [3] ongp@uthm.edu.my; [3] sairajkishore13@gmail.com
* corresponding author

ABSTRACT

Activation functions are essential for deep learning methods to learn and perform complex tasks such as image classification. Rectified Linear Unit (ReLU) has been widely used and become the default activation function across the deep learning community since 2012. Although ReLU has been popular, however, the hard zero property of the ReLU has heavily hindering the negative values from propagating through the network. Consequently, the deep neural network has not been benefited from the negative representations. In this work, an activation function called Flatten-T Swish (FTS) that leverage the benefit of the negative values is proposed. To verify its performance, this study evaluates FTS with ReLU and several recent activation functions. Each activation function is trained using MNIST dataset on five different deep fully connected neural networks (DFNNs) with depth vary from five to eight layers. For a fair evaluation, all DFNNs are using the same configuration settings. Based on the experimental results, FTS with a threshold value, T=-0.20 has the best overall performance. As compared with ReLU, FTS (T=-0.20) improves MNIST classification accuracy by 0.13%, 0.70%, 0.67%, 1.07% and 1.15% on wider 5 layers, slimmer 5 layers, 6 layers, 7 layers and 8 layers DFNNs respectively. Apart from this, the study also noticed that FTS converges twice as fast as ReLU. Although there are other existing activation functions are also evaluated, this study elects ReLU as the baseline activation function.

## 1. Introduction

In deep learning, activation function enables a deep neural model to learn, understand and perform a complicated task by introducing nonlinearity properties into the network. Since the rise of deep learning in 2012, a notable nonsaturated activation function called Rectified Linear Unit (ReLU) [1], [2] has shown its tremendous performance in deep learning [3]. Numerous practical works were done in the past have proven the effectiveness of ReLU across different application domains [4]–[8]. This abrupt paradigm shift in the community is mainly due to two advantages of ReLU. Firstly, the sparsity component in the ReLU. The sparsity arises when *x* < 0. Concisely, ReLU prunes the negative input by outputting zero and retains the positive part [9]. With the sparsity element, ReLU networks are easy to train which resulted in reduces the overall computational cost and substantially expedites the convergence speed. Secondly, ReLU less susceptible to the gradient vanishing problem. Since the derivative of ReLU is 1 at the identity part and 0 otherwise, thus it does not have contractive property as in Sigmoid or Tanh activation functions [10].

Despite the superiority of ReLU, the excessive amount of sparsity element introduced by ReLU could be harmful where it completely prevents the negative values to be propagated through the network. In short, ReLU treats all negative values as unimportant representation. Consequently, deep neural networks have not been benefited from the negative representations. In fact, several studies were done in the past revealing that the negative representation could benefit the network and result in better predictive performance [4], [11]. Since ReLU was used in deep learning, there are several variants of ReLU that allow the negative values to be propagated in the network were introduced. For instance, Leaky ReLU (LReLU) [4], Parametric ReLU (PReLU) [11], Randomized ReLU (RReLU) [9], Exponential linear units (ELU) [10], Gaussian Error Linear Units (GELU) [12] and Scaled Exponential Linear Units (SELU) [13].

Driven by the significance of negative representation, this paper proposes a thresholded ReLU-Swish-like activation function called Flatten-T Swish (FTS), which allows negative values to be propagated in the network and improve overall performance. By looking from the other perspective, this newly activation function co-inherits the properties from ReLU and Swish, furthermore, with a threshold value $T$ attached onto it.

## 2. Method

### 2.1. Flatten-T Swish (FTS) and Rectified Linear Unit (ReLU)

Fig. 1 provides the visualization of FTS (when $T$ at 0.00) and ReLU. From the experiments, the finding shows that: 1) FTS outperforms (classification improvement by 0.13%, 0.70%, 0.67%, 1.07% and 1.15% on wider 5 layers, slimmer 5 layers, 6 layers, 7 layers and 8 layers DFNNs respectively) ReLU consistently as well as other existing activation functions on deep fully connected neural networks (DFNNs) with various depth applied to MNIST dataset classification [14]; 2) FTS converges about 2 times faster than ReLU.

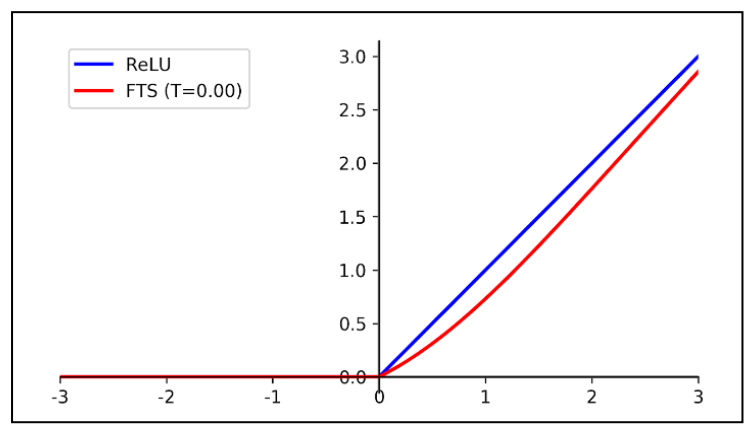

**Fig. 1.** FTS ($T$ = 0.00) vs. ReLU

### 2.2. The Proposed Method: Flatted-T Swish

As mentioned earlier, FTS contains similar properties from both ReLU and Swish, and a threshold $T$ parameter is attached, which could improve the classification accuracy. Mathematically, ReLU is defined as [4]:

$$ReLU(x) = \begin{cases} x, & x \geq 0 \\ 0, & x < 0 \end{cases} \quad (1)$$

ReLU has introduced to tackle the issues such as gradient vanishing/exploding and squashing problems by the Sigmoid activation functions in deep neural networks [15]–[17]. Formally, Sigmoid activation function can be defined as [18]:

$$Sigmoid(x) = \frac{1}{1+e^{-x}} \quad (2)$$

To construct the FTS activation function, this study first amends the original ReLU function by multiplying its linear identity part (when $x \geq 0$) with Sigmoid activation function. Where the idea can be simply expressed by FTS($x$) = ReLU($x$) * Sigmoid($x$) or:

$$FTS(x) = \begin{cases} \frac{x}{1+e^{-x}}, & x \geq 0 \\ 0, & x < 0 \end{cases} \quad (3)$$

With this amendment, this study has noticed that the FTS at $x \geq 0$ has a similar property to a recent activation function introduced by Google Brain called "Swish" [19]. Fig. 2 shows the comparison of the shape of FTS and Swish. Swish has shown its superiority over ReLU on several deep models in image classification and machine translation tasks [19]. However, the derivative of Swish has a large portion of the nonsparse property thus probably trigger higher computational complexity. Meanwhile, FTS retains the hard zero property at the other side as in ReLU which eventually deactivated most of the neurons when during both forward and backward propagation.

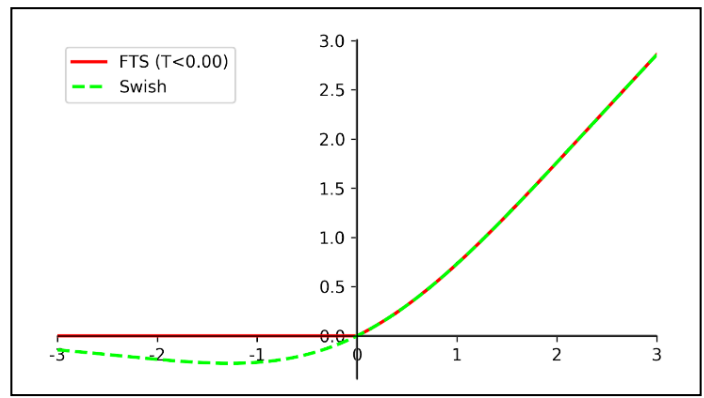

**Fig. 2.** The comparison between FTS (before $T$ is incorporated into the function) and Swish activation functions.

To tackle the ReLU's hard zero problem particularly during the forward propagation, a threshold value $T$ is added to the FTS. This study suggesting that the value for $T$ is set to be less than zero in order to benefit the network with the representations in the negative form. Fig. 3 plots the visualization of the FTS function at $T < 0$. With $T$ added in the function, the function will eventually return all negative values as $T$ when the input value falls at $x < 0$ domain.

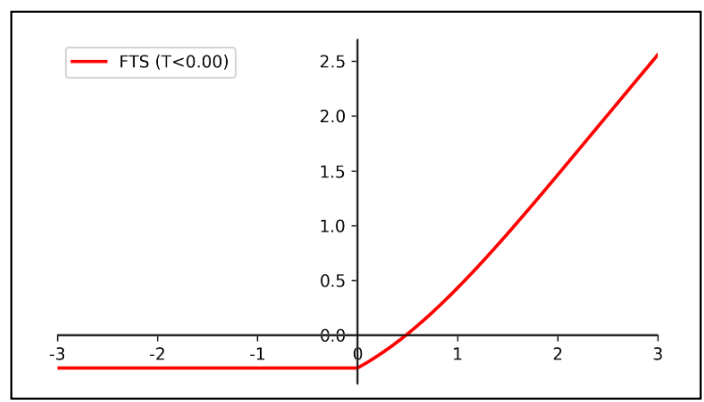

**Fig. 3.** The illustration of FTS at $T < 0$.

Ultimately, FTS with $T$ added is expressed as follows:

$$FTS(x) = \begin{cases} \frac{x}{1+e^{-x}} + T, & x \geq 0 \\ T, & x < 0 \end{cases} \quad (4)$$

Since that the deep neural network is a composition of many differentiable functions [20], therefore, during the backward propagation [21], deep neural network updates its parameters (typically weights and biases) by simply compute its derivative (or gradient). Derivative of a function can be derived by using the chain rule [22]. In the case of FTS function, the chain rule is formulated as:

$$f(x) = g(x).h(x) \tag{5}$$

$$f'(x) = g'(x).h(x) + g(x).h'(x) \tag{6}$$

Then, FTS function is re-denoted as follows:

$$FTS(x) = \begin{cases} f(x), & x \geq 0 \\ T, & x < 0 \end{cases} \tag{7}$$

Sigmoid function is denoted as $\sigma(x)$ in $f(x)$ at the condition when $x \geq 0$. With that, $f(x)$ is expressed as follows:

$$f(x) = x.\sigma(x) + T \tag{8}$$

Since that $T$ is a constant value, its derivative is simply turning to be 0 (similarly, this also applied to the derivative of FTS($x$) during the state where $x < 0$). Therefore, the only term that involves in derivation is $f(x) = x.\sigma(x)$. Its derivative step is listed as follows:

$$f'(x) = 1.\sigma(x) + x(1 - \sigma(x)) \tag{9}$$

$$f'(x) = \sigma(x) + x.\sigma(x) - x.\sigma(x)^2 \tag{10}$$

$$f'(x) = \sigma(x) + f(x) - \sigma(x).f(x) \tag{11}$$

$$f'(x) = \sigma(x) - \sigma(x).f(x) + f(x) \tag{12}$$

$$f'(x) = \sigma(x)(1 - f(x)) + f(x) \tag{13}$$

As a whole, the derivative of the FTS is given by:

$$FTS'(x) = \begin{cases} \sigma(x)(1 - f(x)) + f(x), & x \geq 0 \\ 0, & x < 0 \end{cases} \tag{14}$$

It is worth noting that the derivative of FTS function at the positive part gives similar properties as derived Swish, while the negative part generates similar property as derived. This is clearly indicated that the FTS introduces sparsity only during the backpropagation. The derivative of FTS is shown in Fig. 4, while the difference between derived ReLU and derived FTS can be noticed in Fig. 5.

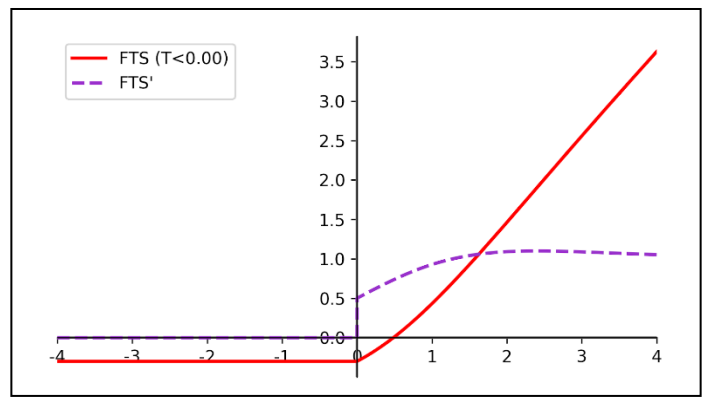

**Fig. 4.** The plot of the FTS function at $T < 0$ and its derivative. The derivative of FTS is denoted as FTS' as shown in the plot label located on the upper-left corner.

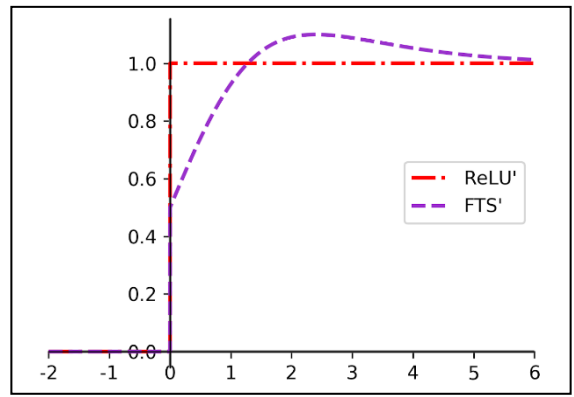

**Fig. 5.** The derived ReLU and derived FTS.

## 3. Results and Discussion

### 3.1. Deep Models and Configuration Settings

The experiments are conducted based on the Python [23] programming language and Tensorflow [24], [25] is used as a computational framework for building the deep models. The entire experiment is executed on Ubuntu 16.04 with a GeForce GTX 1060 6GB Graphics Processing Unit (GPU) to speed up the training.

Since that this study is a preliminary study on this newly propose activation function, therefore the experiment considers deploying the FTS on five different DFNNs with various depth from 5 to 8 layers. The details of the five DFNN architectures are presented in Table 1. The models are trained on 10-classes handwritten digits image dataset known as MNIST [21] which is a commonly used benchmark dataset in various image processing and computer vision experiments. MNIST dataset comprises 60,000 images for training and 10,000 images for testing. Each image is represented by 28 x 28 pixels with the grayscale value ranging from 0 to 255.

For a fair evaluation, similar experiment configurations are used across the deep models. The experiment uses scaled uniform distribution [26] for weight initialization. This method is known as Xavier initialization where commonly used in deep learning [27]–[29]. The mini-batch size is set to be 64 running on vary models with 20 epochs per training. Meanwhile, Stochastic Gradient Descent (SGD) [30] as the optimizer with a learning rate of 0.1 without momentum and weight decay. The dropout rate is set to be 50%. Following the Google Brain in [19], Batch Normalization (BN) [31] is not in used due to some high-level libraries turn off the scale parameter by default on some activation functions.

Table 1. Network architectures of five different DFNNs.

| Network models | Number of hidden layer | Number of neuron in each layer |
|---|---|---|
| DFNN-5a | 5 | 512-512-512-512-10 |
| DFNN-5 | 5 | 256-128-64-32-10 |
| DFNN-6 | 6 | 512-256-128-64-32-10 |
| DFNN-7 | 7 | 784-512-256-128-64-32-10 |
| DFNN-8 | 8 | 1568-784-512-256-128-64-32-10 |

### 3.2. Existing Activation Functions for Comparison

As for the performance comparison, this study compares the FTS with other six commonly used and recent proposed activation functions. Those existing activation functions as well as their details such as the parameter settings are described as follows:

A. ReLU [32]: It was first introduced in 2000 [1] and applied in deep learning models for the first time in 2011 [33]. Since then, it has been chosen as the default activation function in deep learning community. This study treats ReLU as the baseline activation function for performance comparison purpose.

B. Leaky ReLU (LReLU) [4]: LReLU was first proposed in 2013 to address the dying ReLU problem. Where a small positive slope is introduced at $x < 0$ by multiply to as small constant $\alpha = 0.01$. Do note that the experiment also increases the $\alpha$ to 0.25 to see its performance as compared to FTS.

C. Exponential Linear Unit (ELU) [10]: ELU was introduced in 2016 which has shown its superiority in outperforming ReLU in images classification task. In contrast to ReLU, ELU uses exponential property at $x < 0$ to allow the activation to behave slightly like BN which resulted in better generalization and speed up learning.

D. Softplus [34]: Softplus was introduced in 2000 where first applied to model the price of call options. Unlike most of the in-used deep learning activation functions, Softplus is a continuous and smooth function.

E. Swish [19]: Swish was introduced to deep learning particularly in image classification and machine translation tasks by Google Brain team in 2017. In fact, it was similar to Sigmoid-weighted Linear Unit (SiL) [35] function which was used in reinforcement learning. It has the smooth property similar to Softplus. Swish uses parameter $\beta$ (can either be constant or trainable) to control the curvature of the function. However, by following the works in [19], [36], the parameter $\beta$ is fixed to be 1 during the experiment.

F. Flexible ReLU (FReLU) [37]: FReLU is a recently introduced ReLU-like activation function. It has an exactly similar shape as ReLU, but with an additional flexible parameter $b$ to control the function shifted vertically. Since the parameter $b$ is proven to be approximately equal to -0.398 in [37], therefore, the experiment adopts that as the constant for $b$ in FReLU throughout the evaluation process.

Fig. 6 shows the visualization of the activation functions. Although the main objective of the experiment is to evaluate the potential of FTS against the ReLU, yet this study does not rule out the possibility of FTS could also be outperforming other activation functions in this case of implementation.

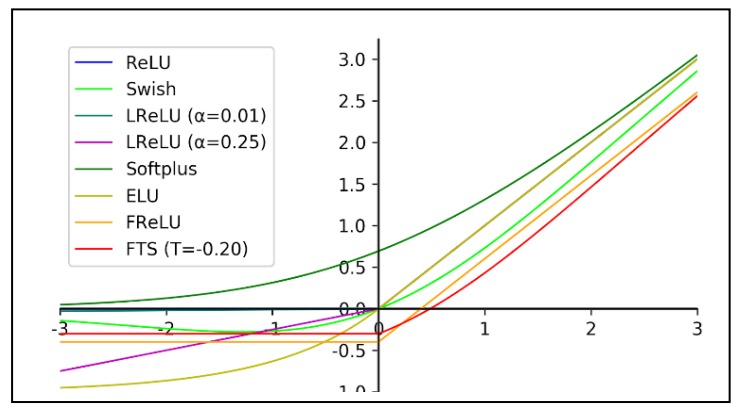

Fig. 6. Visualization of the collective activation functions. Best viewed in color.

### 3.3. FTS with $T = 0.00$

The experiment first evaluates FTS with $T = 0.00$. Notice that the FTS ($T = 0.00$) has exactly similar hard zero property as ReLU that restrains the negative value by outputting zero for any input value that falls within $x < 0$. Yet, another part of the activation is simply the scaled of Sigmoid function which equivalent to the Swish activation at the state where $x \geq 0$. After training the activation functions on DFNN-5a for 20 epochs, the results show that LReLU ($\alpha = 0.01$) has the best achievement (noted with an asterisk*) among the existing activation functions in term of it classification test accuracy. Meanwhile, the experiment also observed that FTS ($T = 0.00$) achieves a slightly better performance recorded as 98.13%, which measured to be 0.02% and 0.07% higher than LReLU ($\alpha = 0.01$) and ReLU respectively. Table 2 shows FTS ($T$ = 0.00) in comparison to existing activation functions in term of mean classification test accuracy.

Table 2. The mean classification test accuracy of activation functions on DFNN-5a with 20 epochs of training.

| Activation function | Test accuracy (%) (Mean of 5 runs) |
|---|---|
| ReLU | 98.06 |
| Swish | 97.97 |
| LReLU ($\alpha$ = 0.01) | 98.11* |
| LReLU ($\alpha$ = 0.25) | 97.64 |
| Softplus | 95.28 |
| ELU | 96.96 |
| FReLU | 98.09 |
| FTS ($T$ = 0.00) | **98.13** |

Apart from that, an additional evaluation is carried out in this section by training the FTS ($T$ = 0.00) and existing activation functions on a relatively smaller 5 layers networks, DFNN-5, hence the results are observed. The result reported in Table 3 shows that FTS ($T$ = 0.00) outperforms ReLU by 0.05%. However, unfortunately, the existing activation functions such as Swish and FReLU turn out to have better performance than FTS ($T$ = 0.00) by a significant margin of 0.55% and 0.50% respectively. Although the result of FTS seems to be less promising at this level, the experiment hypothesizes that by giving a slight margin of negative value for $T$ will improve the overall performance. Therefore, the experiment decreases the $T$ says by 0.05 to allow small negative representation to be captured by the network during the training. As hypothesized, the result shown in Table 4 reveals that FTS ($T$ = -0.05) performs better than FTS ($T$ = 0.00) with an improvement of 0.29%. This revealed that the negative representation could increase the overall network performance. The experiment further explores the FTS to discover a value for $T$ that could generalized well across the models.

Table 3. The mean test accuracy of the activation functions on DFNN-5 with 20 epochs of training.

| Activation function | Test accuracy (%) (Mean of 5 runs) |
|---|---|
| ReLU | 96.96 |
| Swish | **97.65*** |
| LReLU ($\alpha$ = 0.01) | 96.83 |
| LReLU ($\alpha$ = 0.25) | 96.44 |
| Softplus | 94.46 |
| ELU | 96.52 |
| FReLU | 97.6 |
| FTS ($T$ = 0.00) | 97.1 |

Table 4. The mean test accuracy comparison between FTS ($T$ = 0.00) and FTS ($T$ = -0.05) on DFNN-5.

| Activation function | Test accuracy (%) (Mean of 5 runs) |
|---|---|
| FTS ($T$ = 0.00) | 97.1 |
| FTS ($T$ = -0.05) | 97.39 |

### 3.4. A more generalized $T$

It is worth to notice that this experiment does not adopt any method for learning the $T$ parameter or make it as a trainable parameter, though these could be the better approaches to discover a more accurate $T$. These ideas have been put as a part of continuation for this experiment in future work. Based on the previous works [4], [9], [11] that dealing with the ReLU zero hard property, a common trend shows that most of the activation functions that own negative property allow only a small fraction of negative value to be on it. Therefore, this experiment conducts a straightforward approach to discover a generalized $T$ by decreasing it with a small fraction 0.05 on each round of evaluation across the DFNNs.

Table 5 shows the results of FTS with vary of $T$ and existing activation functions. Meanwhile, the numbers in "score" column report the aggregate number of times of each FTS outperforming the best result (noted with an asterisk*) obtained by the existing activation function across the five DFNNs. From the results in Table 5, apparently, FTS with $T$ = -0.20 has the best overall performance where it outperforms existing activation functions in all five evaluations on different models. Hence, a conservative conclusion can be drawn at this stage that -0.20 can probably be considered as a generalized parameter for $T$, at least in this context.

As for the comparison with ReLU baseline, the experiment notice that FTS with $T$ ranging from $0.00 \geq T \geq -0.40$ have basically outperformed ReLU in almost all the cases. By comparing the FTS ($T$ = -0.20) with ReLU in specific, FTS ($T$ = -0.20) improves classification accuracy by 0.13%, 0.70%, 0.67%, 1.07% and 1.15% on DFNN-5a, DFNN-5, DFNN-6, DFNN-7 and DFNN-8 respectively.

Table 5. The test accuracy of FTS with vary of $T$ and existing activation functions. The "score" column aggregate the number of times of each FTS outperforming the best result obtained by the existing activation function.

| Activation function | Test accuracy (%) (Mean of 5 runs) | | | | | Score |
|---|---|---|---|---|---|---|
| | DFNN-5a | DFNN-5 | DFNN-6 | DFNN-7 | DFNN-8 | |
| ReLU | 98.06 | 96.96 | 97.27 | 96.76 | 96.64 | - |
| Swish | 97.97 | 97.65* | 97.85* | 97.77* | 97.64* | - |
| LReLU ($\alpha$ = 0.01) | 98.11* | 96.85 | 97.17 | 96.66 | 97.02 | - |
| LReLU ($\alpha$ = 0.25) | 97.64 | 96.44 | 97.15 | 97.30 | 97.35 | - |
| Softplus | 95.28 | 94.46 | 77.43 | Not converge | Not converge | - |
| ELU | 96.98 | 96.52 | 96.89 | 97.00 | 97.17 | - |
| FReLU | 98.09 | 97.60 | 97.75 | 97.58 | 97.36 | - |
| FTS ($T$=0.00) | **98.13** | 97.10 | 96.96 | 95.77 | 74.65 | 1 |
| FTS ($T$=-0.05) | **98.16** | 97.39 | 97.51 | 97.37 | 97.31 | 1 |
| FTS ($T$=-0.10) | **98.15** | 97.56 | 97.84 | **97.91** | **97.87** | 3 |
| FTS ($T$=-0.15) | **98.19** | 97.61 | 97.84 | **97.87** | **97.74** | 3 |
| FTS ($T$=-0.20) | **98.19** | **97.66** | **97.94** | **97.83** | **97.79** | 5 |
| FTS ($T$=-0.25) | **98.12** | **97.68** | 97.83 | **97.88** | 97.54 | 3 |
| FTS ($T$=-0.30) | 98.10 | **97.75** | 97.82 | **97.85** | 97.44 | 2 |
| FTS ($T$=-0.35) | 98.07 | 97.60 | 97.69 | 97.61 | 97.42 | 0 |
| FTS ($T$=-0.40) | 98.01 | 97.56 | 97.69 | 97.72 | 97.46 | 0 |

Note: The values with asterisk (*) indicate the best results obtained by existing activation functions on respective model. While the values in bold indicate the results of FTS that outperformed the result noted with asterisk (*) on respective model.

### 3.5. Convergence Rate

Apart from evaluating the performance based on the test accuracy, evaluating from the perspective of convergence speed could also telling how well an activation function performs. The experiment trains the DFNN-8 with FTS ($T$ = -0.20) and ReLU using MNIST dataset for 20 epochs. The training and testing accuracy curves of both activation functions are plotted as shown in Fig. 7. Surprisingly, the experiment results show that FTS ($T$ = -0.20) convergence about 2 times faster than ReLU. This again confirmed by the previous works [11], [19], [37] that the activation function that has a slight negative property at $x < 0$ tend to converge much faster than ReLU.

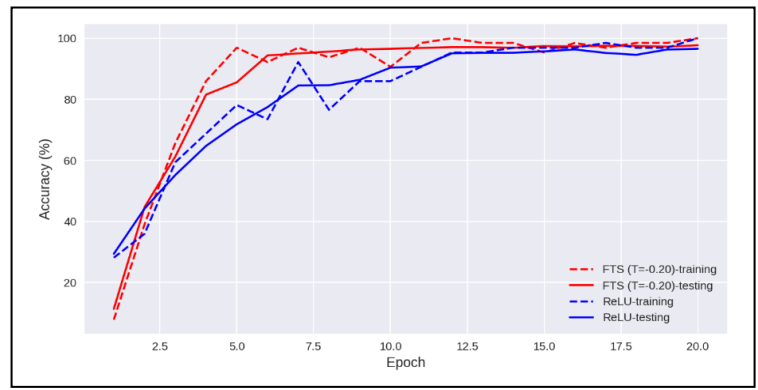

**Fig. 7.** The training and testing curves of FTS ($T$ = -0.20) and ReLU on DFNN-8 network. Best viewed in 0063olor.

## 4. Conclusion

In this paper, a ReLU-Swish-like activation function called Flatten-T-Swish (FTS) is presented. In contrast to ReLU, FTS activation function has a small threshold value $T$ is added to allow negative representations to be flown through the entire network, particularly during the forward propagation. This property enables the network benefits from the negative representations and leads to better predictive performance. Meanwhile, FTS retains the sparsity property during backpropagation where its derivative returns zero at $x < 0$, which is an important element to reduce the computational complexity. The experiment has shown that the FTS particularly with $T$ = -0.20 outperformed other existing activation functions consistently in all five DFNNs with various depth. Specifically, by comparing with ReLU baseline, FTS ($T$ = -0.20) improves MNIST classification accuracy by 0.13%, 0.70%, 0.67%, 1.07% and 1.15% on DFNN-5a, DFNN-5, DFNN-6, DFNN-7 and DFNN-8 respectively. In addition, the experiment also observed that FTS does speed up convergence about 2 times faster than ReLU. As according to the work in [10], [13], [37], this work yet again confirms the importance of negative value in contributing to the overall network performance.

### Acknowledgment

The authors appreciate the efforts of the Office for Research, Innovation, Commercialization and Consultancy Management (ORICC) for providing the Postgraduate Research Grant (GPPS) under Vot U817 and Universiti Tun Hussein Onn Malaysia (UTHM) for supporting this work.